\documentclass[10pt,twocolumn,letterpaper]{article}

\usepackage{cvpr}              

\usepackage[table,xcdraw,dvipdfmx]{xcolor}
\usepackage{graphicx}
\usepackage{amsmath}
\usepackage{amssymb}
\usepackage{booktabs}
\usepackage{comment}
\usepackage{bm}
\usepackage{multirow}
\usepackage{colortbl}
\usepackage{here}

\usepackage[pagebackref,breaklinks,colorlinks]{hyperref}

\usepackage[capitalize]{cleveref}
\crefname{section}{Sec.}{Secs.}
\Crefname{section}{Section}{Sections}
\Crefname{table}{Table}{Tables}
\crefname{table}{Tab.}{Tabs.}

\usepackage{color}

\newcommand{\cSora}[1]{{\color{black}{#1}}}
\newcommand{\cRio}[1]{{\color{black}{#1}}}
\newcommand{\cHiro}[1]{{\color{black}{#1}}}

\def\cvprPaperID{1946} 
\def\confName{CVPR}
\def\confYear{2023}

\begin{document}

\title{
Visual Atoms: Pre-training Vision Transformers with Sinusoidal Waves
}

\author{Sora Takashima$^{1,2}$, Ryo Hayamizu$^{1}$, Nakamasa Inoue$^{1,2}$, Hirokatsu Kataoka$^{1}$, Rio Yokota$^{1,2}$ \vspace{10pt}\\
$^{1}$National Institute of Advanced Industrial Science and Technology (AIST)\\
$^{2}$Tokyo Institute of Technology\\
\footnotesize{\url{https://masora1030.github.io/Visual-Atoms-Pre-training-Vision-Transformers-with-Sinusoidal-Waves/}}
}
\maketitle


\def\figA{
\begin{figure*}
    \centering
    \includegraphics[width=17cm]{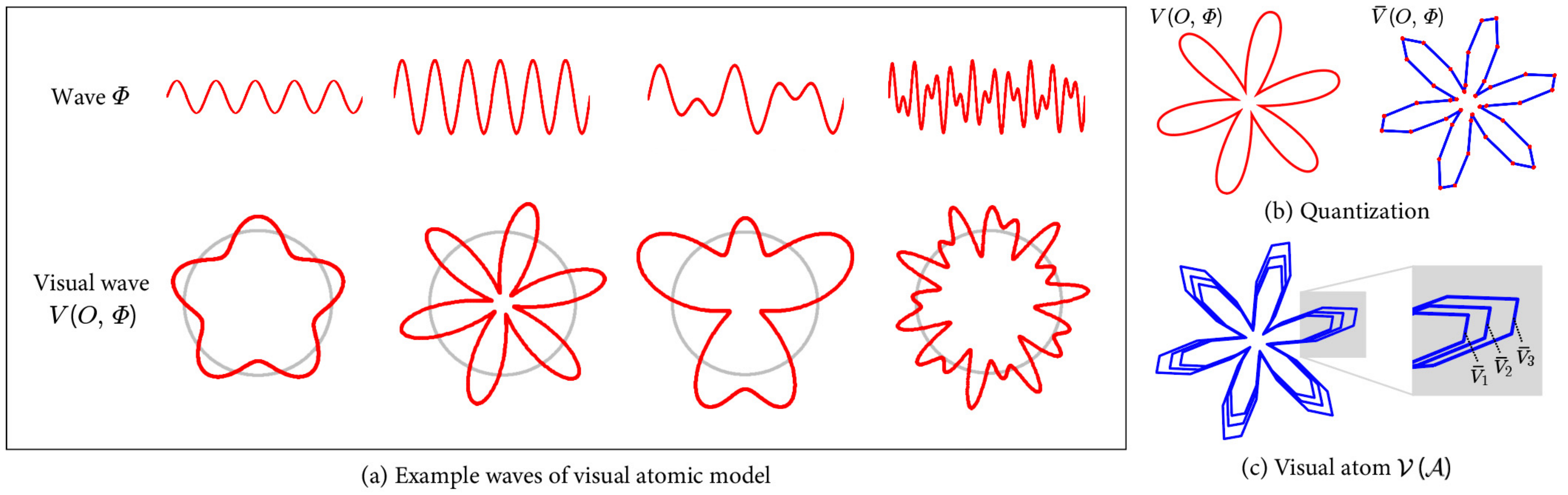}
    \vspace{-8pt}
    \caption{\cSora{{\bf Visual atomic renderer.} (a) De Broglie's atomic model consists of electron orbits $O_{k}$ each with a wave function $\Phi_{k}$ and examples of wave functions $\Phi$ for the visual atomic \cSora{renderer} with their visualization $V(O, \Phi)$. The orbit $O$ is colored gray and the visualized wave $V(O, \Phi)$ is colored in red. (b) Example of quantization from $V(O, \Phi)$ to $\bar{V}(O, \Phi)$. (c) Visual atom $\mathcal{V}(\mathcal{A})$ consisting of visualized quantized waves $\bar{V}_{k} = \bar{V} (O_{k}, \Phi_{k}; q)$.}}
    \vspace{-16pt}
    \label{figA}
\end{figure*}
}

\def\figD{
\begin{figure}[t]
    \centering
    \includegraphics[width=7.0cm]{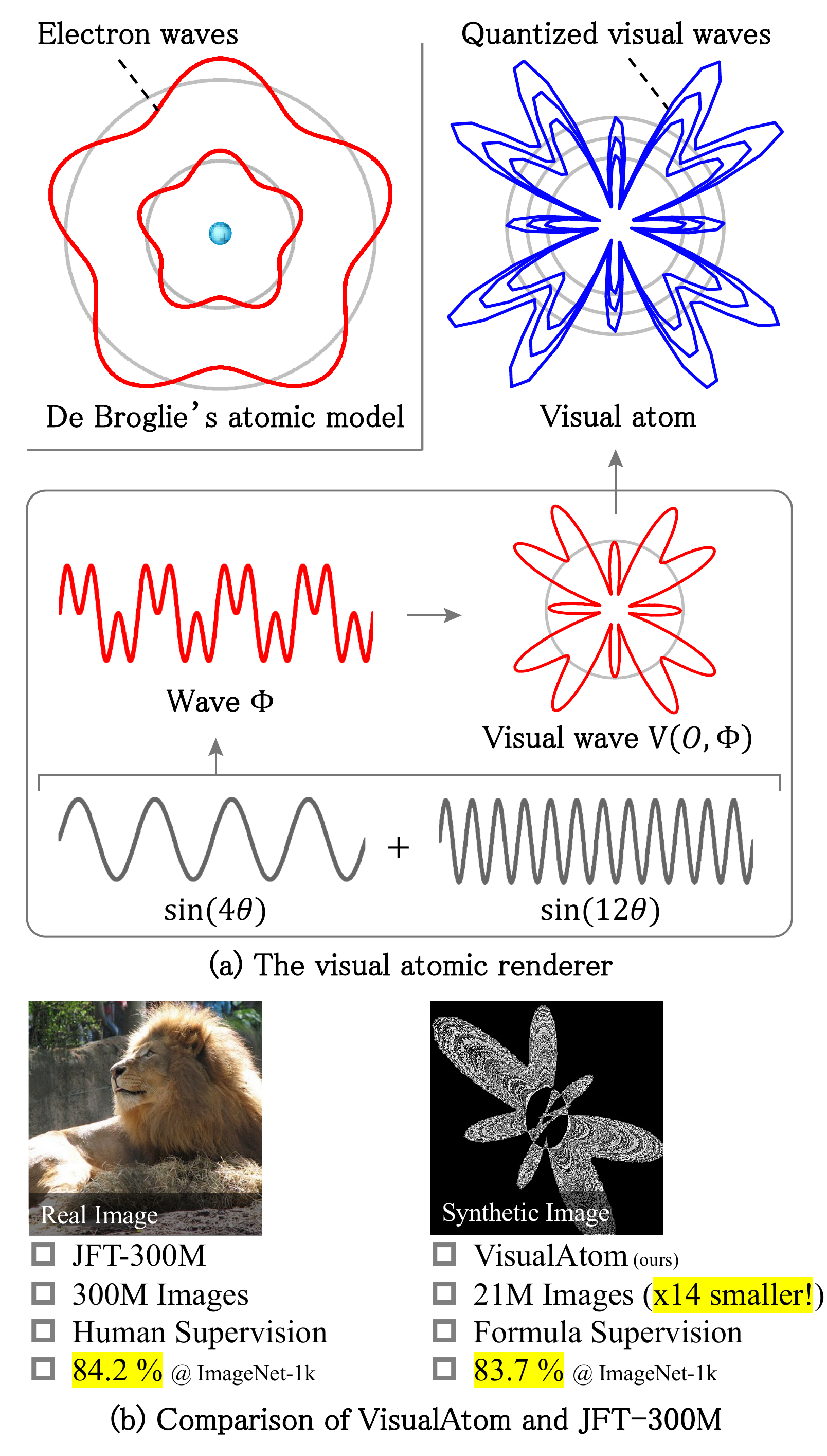}
    \caption{\cSora{{\bf VisualAtom: a new FDSL dataset} (a) Inspired by de Broglie's atomic model, we propose VisualAtom dataset containing shapes from two sinusoidal waves. (b) When fine-tuning on ImageNet-1k, ViT-B pre-trained with VisualAtom achieved the top-1 accuracy close to JFT-300M using about 1/14 images.}}
    \label{figD}
    \vspace{-12pt}
\end{figure}
}

\def\tableA{
\centering
\caption{Parameters of the visual atomic \cSora{renderer}.
Gr: group id by which the parameters are divided into two groups.}
\vspace{-8pt}
\begin{tabular}{cl|c}
\toprule
Gr& Parameter & \cSora{Baseline} Range \\
\midrule
\multirow{6}{*}{1}& Number of orbits $K$ & $\{1, 2, \cdots, 200\}$\\
& Orbit parameters $a_{1}, b_{1}$ & $[1.0, 400.0]$\\
& Orbit interval parameter $c$ & $1.0$ \cSora{(fixed)}\\
& Frequency parameters $n_{i}$ & $\{0, 1, \cdots, 20\}$\\
& Amplitude parameters $\lambda_{i}$ & $0.5$ \cSora{(fixed)}\\
& Quantization parameter $q$ & $\{200, \cdots, 1000\}$\\
& Noise level $\eta$ & $[0.0, 1.0]$\\
\midrule
\multirow{4}{*}{2}&Line-thickness parameter $l_{t}$ & $1$ \cSora{(fixed)}\\
& Line-color parameter $l_{c}$ & $[0.0, 1.0]$\\
& Nucleus position $\bm{p}$ & $[-1.0, 1.0]^{2}$\\
\bottomrule
\end{tabular}
\vspace{-12pt}
\label{tableA}
}

\def\tableC{
\begin{table}[t]
    \vspace{-8pt}
    \centering
\caption{Fine-tuning results on COCO dataset. Best values for each learning type are shown in bold.}
    \vspace{-8pt}
    \begin{tabular}{lccc} \toprule[0.8pt]
        \hspace{-3pt} Pre-training &  COCO Det & COCO Inst Seg \\
         &  AP$_{50}$ / AP / AP$_{75}$ & AP$_{50}$ / AP / AP$_{75}$ \\
        \midrule[0.5pt]
        \hspace{-3pt}Scratch  & 63.7 / 42.2 / 46.1 & 60.7 / 38.5 / 41.3 \\
        \midrule[0.5pt]
         \hspace{-3pt}ImageNet-1k  & 69.2 / 48.2 / 53.0 & 66.6 / 43.1 / 46.5 \\
        \hspace{-3pt}ImageNet-21k & \textbf{70.7} / \textbf{48.8} / \textbf{53.2} & \textbf{67.7} / \textbf{43.6} / \textbf{47.0} \\
        \hspace{-3pt}ExFractalDB-1k\hspace{-16pt} & 69.1 / \textbf{48.0} / \textbf{52.8} & 66.3 / \textbf{42.8} / 45.9 \\
        \hspace{-3pt}ExFractalDB-21k\hspace{-16pt} & \textbf{69.2} / \textbf{48.0} / 52.6 & \textbf{66.4} / \textbf{42.8} / \textbf{46.1} \\
        \hspace{-3pt}RCDB-1k & 68.3 / 47.4 / 51.9 & 65.7 / 42.2 / 45.5 \\
        \hspace{-3pt}RCDB-21k & 67.7 / 46.6 / 51.2 & 64.8 / 41.6 / 44.7 \\
        \midrule[0.5pt]
        \hspace{-3pt}VisualAtom-1k & \cSora{68.8} / \cSora{47.8} / \cSora{52.4} & \cSora{66.0} / \cSora{42.6} / \cSora{45.8} \\ 
        \hspace{-3pt}VisualAtom-21k & 68.3 / 47.4 / 52.3 & 65.3 / 42.2 / 45.4 \\
        \bottomrule[0.8pt]
    \end{tabular}
    \vspace{-8pt}
    \label{tableC}
\end{table}
}

\begin{abstract}
\cRio{Formula-driven supervised learning (FDSL) has been shown to be an effective method for pre-training vision transformers, where ExFractalDB-21k was shown to exceed the pre-training effect of ImageNet-21k.
These studies also indicate that contours mattered more than textures when pre-training vision transformers.
However, the lack of a systematic investigation as to why these contour-oriented synthetic datasets can achieve the same accuracy as real datasets leaves much room for skepticism.
In the present work, we develop a novel methodology based on circular harmonics for systematically investigating the design space of contour-oriented synthetic datasets.
This allows us to efficiently search the optimal range of FDSL parameters and maximize the variety of synthetic images in the dataset, which we found to be a critical factor.
When the resulting new dataset VisualAtom-21k is used for pre-training ViT-Base, the top-1 accuracy reached 83.7\% when fine-tuning on ImageNet-1k.
This is close to the top-1 accuracy (84.2\%) achieved by JFT-300M pre-training, while the number of images is 1/14.
Unlike JFT-300M which is a static dataset, the quality of synthetic datasets will continue to improve, and the current work is a testament to this possibility.
FDSL is also free of the common issues associated with real images, \textit{e.g.} privacy/copyright issues, labeling costs/errors, and ethical biases.
}
\end{abstract}

\section{Introduction}
\label{sec:intro}

Vision transformers \cite{dosovitskiy2021an} have made a significant impact on the entire field of computer vision,
and state of the art models in classification \cite{yu2022coca,wortsman2022modelsoup,zhai2022scaling}, object detection \cite{liu2021swinv2,wang2022beit3}, and segmentation \cite{jain2021semask,chen2022densepred,li2022maskdino,wang2022beit3} are now based on vision transformers.
\cSora{The accuracy of vision transformers exceeds that of convolutional neural networks by a considerable margin when the model is pre-trained on huge datasets, such as JFT-300M \cite{Sun_2017_ICCV}.}
\cSora{However, \cRio{the} JFT-300M dataset contains 300M images and 375M labels. It is impossible to manually label all of these images. Efforts to automatically label such datasets is still not as accurate as manual labeling.}
\figD
\cHiro{Self-supervised learning (SSL) is} increasing in popularity, as datasets do not need to be labeled for this mode of training \cite{jing2021sslsurvey}.
\cRio{Although SSL removes the burden of labeling large datasets, the effort to collect/download, store, and load these large datasets remains a challenge.}

One of the major issues in computer vision is that access to huge datasets, such as JFT-300M/3B~\cite{Sun_2017_ICCV,zhai2022scaling} and IG-3.5B~\cite{mahajan2018exploring}, is limited to certain institutions. \cRio{This} makes it difficult for the rest of the community to build upon, or even reproduce, existing work\cSora{s}.
\cSora{This limitation has prompted the creation of open datasets, such as LAION-5B \cite{schuhmann2022laionb}.}
However, the LAION dataset is not curated, which means that it could contain inappropriate content, or be subject to societal bias and/or privacy/copyright issues \cite{birhane2021multimodal,yang2020towards,yang2022faceobfuscation}.
How to curate such large datasets from the perspective of AI ethics and safety is an open area of research, but, in the meantime, an alternative approach to creating large datasets for computer vision is needed.

Formula-Driven Supervised Learning (FDSL) \cite{Kataoka_2020_ACCV,Kataoka_2022_IJCV} has been proposed as an alternative to supervised learning (SL) and SSL with real images.
The term ``formula-driven'' encompasses a variety of techniques for generating synthetic images from mathematical formulae.
\cRio{The rationale here is that, during the pre-training of vision transformers, feeding such synthetic patterns are sufficient to acquire the necessary visual representations.}
These images include various types of fractals~\cite{Kataoka_2020_ACCV,Kataoka_2022_IJCV,anderson2022improving,nakashima2022can,kataoka2022replacing}, geometric patterns~\cite{kataoka2021formula}, polygons and other basic shapes~\cite{kataoka2022replacing}.
The complexity and smoothness of these shapes can be adjusted along with the brightness, texture, fill-rate, and other factors that affect \cRio{the} rendered image.
Labels can be assigned based on the combination of any of these factors, so a labeled dataset of arbitrary quantity can be generated without human intervention.
Furthermore, there is close-to-zero risk of generating images with ethical implications, such as societal bias or copyright infringement.

\cRio{Another major advantage of synthetic datasets is that the quality of images can be improved continuously, unlike natural datasets which can only be enhanced in quantity.
Therefore, by understanding which properties of synthetic images result in the effective pre-training of vision transformers, we could eventually create a synthetic dataset that can surpass the pre-training effect of JFT-300M.
Nakashima \textit{et al.} \cite{nakashima2022can} used fractal images to pre-train vision transformers and found that the attention maps tend to focus on the contours (outlines) rather than the textures.
Kataoka \textit{et al.} \cite{kataoka2022replacing} verified the importance of contours by creating a new dataset from well-designed polygons, and exceeded the pre-training effect of ImageNet-21k with this dataset that consists of only contours.
These studies indicate that contours are what matter when pre-training vision transformers.
}
\cRio{
However, these studies covered only a limited design space, due to the difficulty of precisely controlling the geometric properties of the contours in each image.}

\cRio{The present study aims to conduct a systematic and thorough investigation of the design space of contour-oriented synthetic images.
We systematically investigate the design space of contours by expressing them as a superposition of sinusoidal waves onto ellipses, as shown in Figure~\ref{figD}\textcolor{red}{a}.
In the same way that a Fourier series can express arbitrary functions, we can express any contour shape with such a superposition of waves onto an ellipse.
Such geometrical concepts have appeared in classical physics, \textit{e.g.} de Broglie's atomic model~\cite{debroglie1924wave}.
Therefore, we name this new dataset ``VisualAtom", and the method to generate the images ``visual atomic renderer".}
\cRio{The visual atomic renderer allows us to exhaustively cover the design space of contour-oriented synthetic images, by systematically varying the frequency, amplitude, and phase of each orbit, along with the number of orbits and degree of quantization the orbits.
We vary the range of these parameters to generate datasets with different variety of images.
We found that variety of contour shapes is a crucial factor for achieving a superior pre-training effect.
\cSora{
Our resulting dataset was able to nearly match the pre-training effect of JFT-300M when fine-tuned on ImageNet-1k, while using only 21M images.}}
\cHiro{We summarize the contributions as follows:}

\cRio{
\noindent \cHiro{\textbf{\underline{Investigative contribution (Figure~\ref{figD}\textcolor{red}{a}):}}} We propose a novel methodology based on circular harmonics that allows us to systematically investigate the design space of contour-oriented synthetic datasets.
Identifying the optimal range of frequency, amplitqude, and quantization of the contours lead to the creation of a novel synthetic dataset VisualAtom with unprecedented pre-training effect on vision transformers.
}

\noindent \cHiro{\textbf{\underline{Experimental contribution (Figure~\ref{figD}\textcolor{red}{b}):}}} We show that pre-training ViT-B with \cSora{VisualAtom} can achieve comparable accuracy to pre-training on JFT-300M,da when evaluated on ImageNet-1k fine-tuning. Notably, the number of images used to achieve this level of accuracy was approximately 1/14 of JFT-300M. We also show that VisualAtom outperforms existing state-of-the-art FDSL methods.

\noindent \cHiro{\textbf{\underline{Ethical contribution:}}} We will release the synthesized image dataset, pre-trained models, and the code to generate the dataset. This will also allow users with limited internet bandwidth to generate the dataset locally. Moreover, the dataset and model will be released publicly as a commercially available license and not limited to educational or academic usage.

\section{Related work}

\subsection{Vision Transformers \cHiro{and} Large-scale Datasets}

Since the first ViT paper was published by Dosovitskiy et al.~\cite{dosovitskiy2021an}, there have been numerous extensions and improvements to vision transformers.
During the same time, there have also been various improvements to SL, SSL, and FDSL.
In this section, we will introduce some of the most relevant work in these fields, with the goal of clarifying the contribution of the present work.

\cSora{
ViT~\cite{dosovitskiy2021an} was the first successful attempt to use transformers~\cite{vaswani2017attention} in computer vision, with only minor modifications made to the original architecture.
}
One of the main findings of their study was that when ViT is pre-trained on larger datasets, such as ImageNet-21k and JFT-300M, the accuracy of downstream-tasks scales with the size of datasets \cSora{and models}.
There have also been efforts to use SSL for pre-training vision transformers, e.g. 
DINO~\cite{caron2021emerging}, BEiT~\cite{bao2021beit}, MAE~\cite{he2022masked}.
However, the highest accuracy for ImageNet-1k fine-tuning is still obtained through vision transformers with supervised pre-training using JFT-3B~\cite{zhai2022scaling}.

The largest datasets, such as JFT-300M/3B/4B~\cite{Sun_2017_ICCV,zhai2022scaling}, Instagram-3.5B~\cite{mahajan2018exploring}, and YFCC (Yahoo! Flickr Creative Commons)\footnote{The full version has not yet been released, but a copy of YFCC-15M is available.}~\cite{thomee2016yfcc100m} are all excellent candidates for pre-training vision transformers.
However, access to these datasets is limited to a selected group of researchers.
To address this issue, the LAION-400M/5B dataset~\cite{schuhmann2021laion, schuhmann2022laionb} has been made public under a permissive license that allows commercial use.
\cSora{LAION}-5B contains 5.85 billion CLIP-filtered image-text pairs, which can be used to train large vision+language models.
However, as the LAION dataset is also not curated, its use is subject to the same concerns associated with other large datasets (i.e., the existence of societal bias and privacy and copyright concerns)~\cite{birhane2021multimodal}.
These ethical concerns associated with large datasets are a serious problem.
\cSora{
For example, one of the most widely used datasets in computer vision -- ImageNet~\cite{deng2009imagenet}, contains images that violate privacy protection laws and biased samples, can lead to unfairness in models trained on the dataset ~\cite{yang2020towards}.
}
Furthermore, public releases of datasets such as 80M Tiny Images~\cite{torralba200880} have actually been suspended due to discriminatory labels in the dataset~\cite{birhane2021large}.

SSL can avoid issues related to discriminatory labels, and also frees us from the tedious task of labeling these huge datasets.
However, SSL has not yet achieved the same level of accuracy as SL.
\cRio{
Moreover, SSL still relies on natural images and shares the limitations of large datasets \textit{e.g.} societal bias, privacy and copyright.
}

\subsection{Formula-Driven Supervised Learning}

An alternative method for pre-training vision transformers has been proposed, where synthetic images generated from mathematical formulae are used to perform formula-driven supervised learning (FDSL)~\cite{Kataoka_2020_ACCV, nakashima2022can}.
Several follow-up studies on FDSL have been published that extend the application of this approach and that increase the quality of the synthetic datasets~\cite{anderson2022improving, inoue2021initialization, kataoka2021formula, kataoka2022replacing, yamada2022point}
FDSL has been evolving rapidly, and the quality of images has improved to the point where it now surpasses the level of pre-training that can be achieved using ImageNet-21k~\cite{kataoka2022replacing}.
Further, FDSL has considerable potential because it produces a \textit{dynamic} dataset that can be continuously improved, unlike \textit{static} datasets such as JFT and LAION in which only the quantity of the dataset can be increased.
Another advantage of FDSL is that it is entirely free of the common ethical issues associated with natural images, such as societal bias, privacy, and copyright infringement.
FDSL also does not require the downloading of an enormous number of images from a host website, because the same synthetic images can be generated locally by anyone that has the formula.
These advantages make it a worthwhile endeavor to take the current FDSL that can match ImageNet-21k, and improve its quality to the point where it can match JFT-300M.
\cSora{
However, previous FDSL methods have several outstanding issues: 
1) difficult to control the variety of the contour outline of the synthetic images, 
2) lack of smoothness of the contour, 
3) unclear why and how to generate better synthetic images for pre-training.
\cRio{
Therefore, an exhaustive investigation regarding the shape of contours in synthetic images is required.}
}

\section{Method}

\figA

Conceptually inspired by de Broglie's atomic model in the field of quantum physics, we propose the {\it visual atomic \cSora{renderer}} which can be used to generate synthetic images for learning visual representations in computer vision tasks.
Based on the idea that the various object shapes needed in pre-training can be obtained from combinations of smooth wave functions, we present our \cSora{method} in the following in two steps.
First, we define the visual atomic \cSora{renderer} that \cSora{produces} 2D visualization of atoms.
Second, we present a pre-training method using the \cSora{VisualAtom, which is synthetic image dataset generated by our proposed the visual atomic renderer}.
\cSora{
The visual atomic renderer can assign many variations to the contour outlines and control these variations by changing the parameter ranges.
}

\subsection{Visual Atomic \cSora{Renderer}}

The visual atomic \cSora{renderer} $\mathcal{V}$ visualizes an atom $\mathcal{A}$ as an image.
More specifically, the visual atomic \cSora{renderer produces and renders a {\it visual atom} $\mathcal{V}(\mathcal{A}) \subset \mathbb{R}^{2}$ from an atom $\mathcal{A}$.}
Here, we define $\mathcal{V}$ by parameterizing each component of an atom.

\noindent{\bf Parameterized orbits.}
We define the orbit $O_{k}$ by an ellipse with two parameters $(a_{k}, b_{k}) \in \mathbb{R}^{2}$ in a 2D Euclidean \cHiro{space}: 
\begin{align}
O_{k} = \{(x,y)^{\top}: F_{k}(x,y) = 0\} \subset \mathbb{R}^{2},
\end{align}
where $F_{k}$ is the implicit function of the ellipse given by
\begin{align}
F_{k}(x, y) = \frac{x^{2}}{a_{k}^{2}} + \frac{y^{2}}{b_{k}^{2}} - 1.
\end{align}
Notably, $O_{k}$ has the following parametric representation:
\begin{align}
\label{eq:pramrep} 
O_{k} = \left\{
\left(
\begin{matrix}
a_{k} \cos \theta \\
b_{k} \sin \theta 
\end{matrix}
\right): 0 \leq \theta < 2 \pi
\right\}.
\end{align}
where, $\theta$ is an auxiliary parameter. To reduce the number of parameters when making a sequence of orbits, we introduce the interval parameter $c$ and recursively determine the parameter values as $(a_{k+1}, b_{k+1}) = (a_{k} + c, b_{k} + c)$ given randomly sampled $(a_{1}, b_{1})$.

\noindent{\bf Parameterized waves.}
The simplest definition of the wave function $\Phi_{k}$ can be given by a single sinusoidal wave:
\begin{align}
\Phi_{k}(\theta) = \lambda \sin (n \theta),
\end{align}
where, $n \in \mathbb{N}$ is a frequency parameter and $\lambda \in \mathbb{R}$ is an amplitude parameter.
Note that $n$ needs to be a natural number in order to satisfy the condition $\Phi_{k}(\theta) = \Phi_{k}(\theta + 2 \pi).\hspace{10pt}$

To have a sufficient variety of waves, the visual atomic \cSora{renderer} assumes that each orbit has a mixture of two sinusoidal waves and a small noise as follows:
\begin{align}
\label{wavedef}
\Phi_{k}(\theta) = \lambda_{1} \sin (n_{1} \theta) + \lambda_{2} \sin (n_{2} \theta) + \eta \epsilon(\theta),
\end{align}
where, $n_{1}, n_{2} \in \mathbb{N}$ are frequency parameters,
$\lambda_{1}, \lambda_{2} \in \mathbb{R}$ are amplitude parameters, $\eta \in \mathbb{R}$ is a noise-level parameter, and $\epsilon$ is a noise function such that $\epsilon(\theta) = \epsilon(\theta + 2 \pi)$.
Examples of wave functions (w/o noise) are shown in Figure~\ref{figA}\textcolor{red}{a}. This simple definition using sinusoidal waves will provide diverse shapes that help \cHiro{ViT} to learn visual representations.

\noindent {\bf Visual waves.}
Given an orbit $O_{k}$ with a wave function $\Phi_{k}$,
we define the visual wave $V(O_{k}, \Phi_{k}) \subset \mathbb{R}^{2}$ by multiplying the wave function by the parametric representation of the orbit in Eq.~(\ref{eq:pramrep}) as follows:
\begin{align}
V(O_{k}, \Phi_{k}) = \left\{
\Phi_{k}^{\star}(\theta)
: 0 \leq \theta < 2 \pi
\right\},
\end{align}
where,
\begin{align}
\Phi_{k}^{\star}(\theta) = 
\left(
\begin{matrix}
a_{k} ( \Phi_{k}(\theta) + 1) \cos \theta \\
b_{k} ( \Phi_{k}(\theta) + 1) \sin \theta 
\end{matrix}
\right) \in \mathbb{R}^{2}.
\end{align}
Examples of visual waves are shown in Figure~\ref{figA}\textcolor{red}{a}.
Interestingly, we see that some of them look like the shapes of natural objects, such as a star, a flower, a bird, and an amoeba.

Further, to incorporate sharp corners, we introduce a quantization parameter $q$, which quantizes $\Theta = [0, 2\pi)$ into a discrete set of points $\bar{\Theta} = \{\theta_{i} : i = 0, 1, 2, \cdots q-1\}$, where $\theta_{i} = 2 \pi i / q$.
We define the quantized visual \cHiro{wave:} 
\begin{align}
\label{eq:quantized_wave}
\bar{V}(O_{k}, \Phi_{k}; q) = \bigcup_{i=0}^{q-1} \ell(\bm{x}_{i}, \bm{x}_{i+1}),
\end{align}
where, $\bm{x}_{i} = \Phi_{k}^{\star}(\theta_{i})$ is the $i$-th point on the visual wave and
\begin{align}
\ell(\bm{x}_{i}, \bm{x}_{i+1}) = \{ \alpha \bm{x}_{i} + (1-\alpha) \bm{x}_{i+1} : 0 \leq \alpha \leq 1\}
\end{align}
is the line between the two points ($\bm{x}_{i}$ and $\bm{x}_{i+1}$).
This procedure is illustrated in Figure~\ref{figA}\textcolor{red}{b}.

Finally, given an atom $\mathcal{A}$, we define a visual atom $\mathcal{V}(\mathcal{A})$ by taking the union set of quantized visual waves as follows:
\begin{align}
\label{eq:visual_atom}
\mathcal{V}(\mathcal{A}) = \bigcup_{k=1}^{K} \bar{V}(O_{k}, \Phi_{k}; q).
\end{align}
An example of the visual atom is shown in Figure~\ref{figA}\textcolor{red}{c}.


\noindent {\bf Rendering.}
From Eqs.~(\ref{eq:quantized_wave}) and (\ref{eq:visual_atom}), a visual atom $\mathcal{V}(\mathcal{A})$ can be viewed as a set of lines. 
Thus, we render the lines to synthesize the image.
For rendering, there are two parameters: line thickness $l_{t}$ and line color $l_{c} \in [0.0, 1.0]$ in gray scale.
The line thickness is fixed to $l_{t} = 1$ pixel as the baseline.
The color is randomly selected for each orbit.
Finally, the nucleus (the center of the visual atom) is parameterized by a 2D vector $\bm{p} \in \mathbb{R}^{2}$ by which the rendering position is determined.
All of the lines are rendered on a black background.

\noindent {\bf \cSora{Compiling VisualAtom}.}
We create a pre-training dataset $D$, which consists of pairs $(I, y) \in D$ of an image $I$ and a class label $y \in \{1, 2, \cdots, C\}$, by the following two steps.
First, $C$ independent sets of parameters are randomly sampled to determine the classes of the visual atoms.
Here, we divide the parameters into two groups, as shown in Table~\ref{tableA}.
For each class, the parameters in the first group are randomly sampled and fixed.
Second, $N$ images are generated for each class. Here, the parameters of the second group are randomly selected for each image, as this increases the intra-class variability.
Finally, the training dataset consists of $|D| = N C$ images. \cSora{We names this synthesized image dataset ``VisualAtom''.}
The cross-entropy loss over $D$ is minimized during the pre-training phase.


\def\tableBaseline{
    \begin{center}
    \caption{Comparison of fine-tuning accuracy on four datasets: CIFAR10 (C10), CIFAR100 (C100), ImageNet-100 (IN100) and ImageNet-1k (IN-1k).
    Cross entropy loss is used for SL.
    DINO is used for SSL. ViT-T is used for all experiments.}
    \vspace{-8pt}
    \begin{tabular}{lcccc} \toprule[0.8pt]
        Pre-training & C10 & C100 & IN100 & IN-1k \\\midrule[0.5pt]
        Scratch & 78.3 & 57.7 & 73.2 & 72.6 \\ \midrule[0.5pt]
        ImageNet-1k (SL) & 98.0 & 85.5 & - & - \\
        ImageNet-1k (SSL) & 97.7 & 82.4 & 89.0 & - \\
        \midrule[0.5pt]
        ExFractalDB-1k & 97.2 & 81.8 & 88.1 & 73.0 \\
        RCDB-1k & 97.0 & 82.2 & 88.5 & 73.1 \\
        \rowcolor[gray]{0.8} VisualAtom-1k (ours) & \textbf{97.6} & \textbf{84.9} & \textbf{90.3} & \textbf{74.2} \\
        \bottomrule[0.8pt]
        \multicolumn{5}{l}{
            \begin{minipage}{\hsize}
                \raggedleft
                \vspace{2pt}
                \includegraphics[width=\hsize,clip]{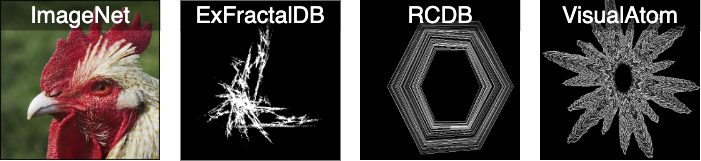}
            \end{minipage}
        } \\
        \bottomrule[0.8pt]
    \end{tabular}
    \label{tab:VisualAtom-1k_baseline}
    \end{center}
    \vspace{-16pt}
}

\def\tabularFrequency{
    \begin{center}
    \caption{Accuracy when varying the range of frequency parameters $n_{1}, n_{2}$.
    }
    \vspace{-8pt}
    \begin{tabular}{cc|ccc} \toprule[0.8pt]
        \multicolumn{2}{c|}{\hspace{-5pt}Range of $n,m$\hspace{-3pt}} & \multirow{2}{*}{C10} & \multirow{2}{*}{C100} & \multirow{2}{*}{\hspace{-3pt}IN100\hspace{-3pt}}\\
        min & max & & &
        \\\midrule[0.5pt]
        \rowcolor[gray]{0.8} 0&20 & {\bf 97.6} & {\bf84.9} & {\bf 90.3} \\
        0&40 & 97.1 & 84.3 & 89.5 \\
        0&60 & 97.3 & 84.1 & 89.1 \\
        2&20 & {\bf 97.6} & 84.8 & {\bf 90.3} \\
        10&20 & 97.5 & 84.4 & 89.8 \\
        20&20 & 97.3 & 83.6 & 89.6 \\
        \bottomrule[0.8pt]
        \multicolumn{5}{l}{
            \begin{minipage}{\hsize}
                \raggedleft
                \vspace{2pt}
                \includegraphics[width=\hsize,clip]{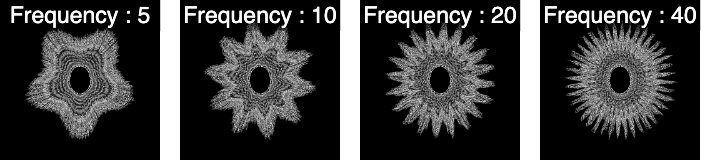}
            \end{minipage}
        }\\
        \bottomrule[0.8pt]
    \end{tabular}
    \label{tab:VisualAtom-1k_nami_change}
    \end{center}
}

\def\tabularQuantization{
    \begin{center}
    \vspace{11pt}
    \caption{Accuracy when varying the range of quantization parameter $q$.
    }
    \vspace{-8pt}
    \begin{tabular}{cc|ccc} \toprule[0.8pt]
        \multicolumn{2}{c|}{Range of $q$} & \multirow{2}{*}{C10} & \multirow{2}{*}{\hspace{-2pt}C100\hspace{-2pt}} & \multirow{2}{*}{\hspace{-3pt}IN100\hspace{-3pt}} \\
        min & max & & &\\
        \midrule[0.5pt]
        \rowcolor[gray]{0.8} 200 & 1,000 & {\bf 97.6} & 84.9 & \textbf{90.3}\\
        800 & 1,000 & 97.4 & {\bf 85.1} & 89.9\\
        3 & 200 & 97.3 & 84.6 & 89.7\\
        3 & 500 & 97.3 & 84.9 & 90.1\\
        3 & 1,000 & 97.4 & 85.0 & 90.1\\
        \bottomrule[0.8pt]
        \multicolumn{5}{l}{
            \begin{minipage}{\hsize}
                \raggedleft
                \vspace{2pt}
                \includegraphics[width=\hsize,clip]{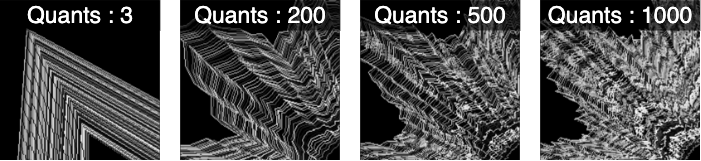}
            \end{minipage}
        }\\
        \bottomrule[0.8pt]
    \end{tabular}
    \label{tab:RCDB_and_VisualAtom-1k_vertex_change}
    \end{center}
}

\def\tabularAmplitude{
    \begin{center}
    \caption{Accuracy when varying the range of amplitude parameters $\lambda_{1}, \lambda_{2}$.}
    \vspace{-8pt}
    \begin{tabular}{cc|ccc} \toprule[0.8pt]
        \multicolumn{2}{c|}{\hspace{-5pt}Range of $\lambda_{1}, \lambda_{2}$\hspace{-3pt}} & \multirow{2}{*}{C10} & \multirow{2}{*}{C100} & \multirow{2}{*}{IN100}\\
        min & max & & &
        \\\midrule[0.5pt]
        \rowcolor[gray]{0.8} 0.5 & 0.5 & {\bf 97.6} & {\bf84.9} & {\bf 90.3} \\
        0.0 & 0.5 & 97.4 & \textbf{84.9} & 89.5 \\
        1.0 & 1.0 & 97.3 & 84.4 & 89.6 \\
        0.5 & 1.0 & 97.4 & 84.5 & 89.6 \\
        0.0 & 1.0 & 97.4 & 84.4 & 89.8 \\
        \bottomrule[0.8pt]
        \multicolumn{5}{c}{
            \begin{minipage}{0.7\hsize}
                \centering
                \vspace{2pt}
                \includegraphics[width=0.7\hsize,clip]{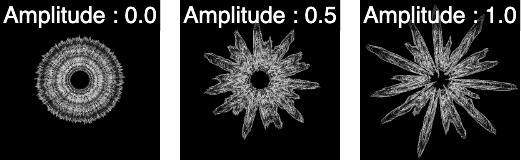}
            \end{minipage}
        }\\
        \bottomrule[0.8pt]
    \end{tabular}
    \label{tab:VisualAtom-1k_amplitude}
    \end{center}
    \vspace{-24pt}
}

\def\tabularNumorbits{
    \begin{center}
    \vspace{13pt}
    \caption{Accuracy when varying the range of number of orbits $K$.}
    \vspace{-8pt}
    \begin{tabular}{cc|ccc} \toprule[0.8pt]
        \multicolumn{2}{c|}{\hspace{-5pt}Range of $K$\hspace{-3pt}} & C10 & C100 & IN100\\
        min & max & & &\\
        \midrule[0.5pt]
        \rowcolor[gray]{0.8} 1 & 200 & {\bf 97.6} & \textbf{84.9} & {\bf 90.3} \\
        20 & 200 & 97.5 & 84.7 & 89.9 \\
        100 & 200 & 97.5 & 84.5 & 89.8 \\
        200 & 200 & 97.5 & 84.3 & 89.4 \\
        \bottomrule[0.8pt]
        \multicolumn{5}{l}{
            \begin{minipage}{\hsize}
                \raggedleft
                \vspace{2pt}
                \includegraphics[width=\hsize,clip]{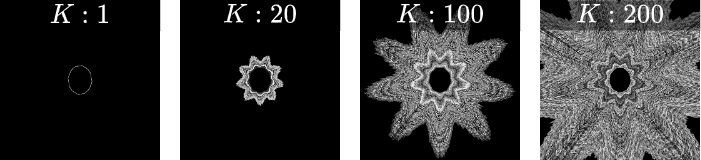}
            \end{minipage}
        }\\
        \bottomrule[0.8pt]
    \end{tabular}
    \label{tab:VisualAtom-1k_number_of_orbits}
    \end{center}
    \vspace{-20pt}
}

\section{Experiments}
\cHiro{We} conduct experiments to validate the effectiveness of \cSora{VisualAtom} for pre-training vision transformers. 
\cSora{
We first explore the relationship between various contour features and the pre-training effect, \cRio{by} changing the parameters of the visual atomic renderer such as frequency in Section~\ref{Pre-training-experiments-on-contours}.
}
We then examine the fine-tuning performance in comparison with state-of-the-art pre-training methods, including supervised and self-supervised pre-training on ImageNet-21k and JFT-300M in Section~\ref{sec:comparision}.
Finally, we discuss considerations and limitations of \cSora{our} method in Section~\ref{sec:discussion}.

\subsection{Pre-training with \cSora{VisualAtom}}
\label{Pre-training-experiments-on-contours}

We first validate the effectiveness of \cSora{VisualAtom} by comparing it with ExFractalDB-1k/RCDB-1k \cite{kataoka2022replacing}. Here, both ExFractalDB-1k/RCDB-1k are state-of-the-art datasets for FDSL that each comprise one million images of well-designed fractals/polygons.
For a fair comparison with ExFractalDB-1k/RCDB-1k in terms of the number of images, we compiled VisualAtom-1k, a dataset comprised of one million images of visual atoms, where the number of classes is $C=1,000$ and
the number of images per class is $N=1,000$.
We also report results when using \cHiro{SL and SSL} on ImageNet-1k~\cite{deng2009imagenet}.
\cSora{Pre-training and fine-tuning of ViT were done with the same hyper-parameters as \cHiro{DeiT}~\cite{pmlr-v139-touvron21a}.}
\begin{table}[t]
\tableA
\end{table}
\begin{table}[h]
\tableBaseline
\vspace{-12pt}
\end{table}

\begin{table*}[htbp]
\begin{minipage}{0.3\hsize}
\tabularFrequency
\end{minipage}
\hspace{12pt}
\begin{minipage}{0.3\hsize}
\tabularNumorbits
\end{minipage}
\hspace{12pt}
\begin{minipage}{0.3\hsize}
\tabularQuantization
\end{minipage}
\vspace{-15pt}
\end{table*}
Table~\ref{tab:VisualAtom-1k_baseline} compares the \cHiro{accuracy}, using the following four datasets: CIFAR10 (C10)~\cite{krizhevsky2009learning}, CIFAR100 (C100)~\cite{krizhevsky2009learning}, ImageNet-100 (IN100)\footnote{This is a subset of ImageNet with 100 object categories.}, and ImageNet-1k (IN-1k).
The findings showed that the proposed VisualAtom-1k outperforms RCDB-1k on all of the four fine-tuning datasets.
Notably, we obtained 2.7 point improvement on the C100 dataset (VisualAtom-1k 84.9 v.s. RCDB-1k 82.2).
In addition, the performance obtained with VisualAtom-1k was comparable to, or better than, SSL (DINO).
Overall, the results confirmed that VisualAtom-1k achieves state-of-the-art performance when used with FDSL.

\cRio{
We then conducted experiments to analyze the effects of individual parameters of the visual atomic renderer, and discuss the importance of each contour feature.
}
The baseline ranges from which parameter values were sampled are summarized in Table~\ref{tableA}.

\noindent {\textbf{Frequency Parameters (Table~\ref{tab:VisualAtom-1k_nami_change}).}}
\label{Contour-vibrations}
The two frequency parameters $n_{1}$ and $n_{2}$ in Eq.~(\ref{wavedef}) are the most important parameters because they determine the overall shape of visual atoms.
Table~\ref{tab:VisualAtom-1k_nami_change} shows the results of experiments conducted using various frequency ranges.
Because $n_{1}$ and $n_{2}$ need to be natural numbers,
we varied the minimum and maximum numbers for them.
For example, the setting $[\text{min}, \text{max}] = [0, 20]$ at the first row indicates that $n_{1}$ and $n_{2}$ are randomly sampled from $\{0, 1, \cdots, 20\}$.

The results show that the best setting is $[0, 20]$ and that VisualAtom-1k outperforms RCDB-1k at all settings.
Increasing the maximum number in the range did not improve the pre-training effect because the frequency of the sinusoidal waves became too high.
The value of 20 was considered to be a reasonable maximum frequency for pre-training with an image size of 224 by 224 [pixel].
In addition, the fine-tuning accuracy was tested when the maximum frequency is fixed at 20, and the minimum frequency is increased gradually from 0 to 20 to narrow the frequency range.
The results showed that the fine-tuning accuracy declined for all downstream-tasks as the frequency range was narrowed. \cRio{We speculate that this decline in accuracy may be caused by the decrease in category variations determined by the combination of the two frequency parameters.}

\noindent {\cSora{\textbf{Number of Orbits Parameter (Table~\ref{tab:VisualAtom-1k_number_of_orbits}).}}}
\label{Contour-nums}
The parameter $K$ defines the number of orbits on one-image. $K$ is sampled uniformly per category from pre-defined natural number range or is fixed at a pre-defined natural number. \cRio{Table~\ref{tab:VisualAtom-1k_number_of_orbits} shows the results when varying the range of $K$.}

Using a maximum $K$ of 200, we found that the fine-tuning accuracy tends to increase for almost all downstream-tasks at the lower minimum $K$ and the longer the length of the range.
\cSora{
This result indicates that a sufficiently \cRio{large} variation of $K$ per category improves the pre-training effect.
This finding supports the hypothesis that \cRio{the} pre-training dataset should contain categories of various size object shapes.
}

\noindent {\textbf{Quantization Parameter (Table~\ref{tab:RCDB_and_VisualAtom-1k_vertex_change}).}}
\label{Contour-vertices}

\cSora{
The quantization parameter $q$ \cRio{controls the smoothness of the contours}.
As shown in Table~\ref{tab:RCDB_and_VisualAtom-1k_vertex_change}, we did not observe \cRio{any} degradation of accuracy even \cRio{for} the smallest range of $[800, 1000]$.
On the other hand, \cRio{when $q$ is in} the range of $[3, 200]$, we observed \cRio{a drop in} accuracy.
Furthermore, we observed that the accuracy improved as the minimum $q$ was fixed to 1 and the maximum $q$ was increased to $500$ or $1000$.

\cRio{These results indicate that the quantization parameter $q$ needs to be above a certain value, for better representation and smoothness of the contours.}
This finding supports our hypothesis that the various object shapes that are needed in pre-training can be obtained from smooth wave functions.
}

\noindent {\cSora{\textbf{Amplitude Parameters (Table~\ref{tab:VisualAtom-1k_amplitude}).}}}
\label{Orbits-amplitudes}
\begin{table}[t]
\tabularAmplitude
\end{table}
\cSora{The two amplitude parameters, $\lambda_{1}$ and $\lambda_{2}$ define the amplitude of two separate sinusoidal waves added on the orbit.
$\lambda_{1}$ and $\lambda_{2}$ are sampled uniformly per category from a pre-defined non-negative decimal range or fixed at a pre-defined decimal. Table~\ref{tab:VisualAtom-1k_amplitude} shows the results of experiments with various amplitude ranges.

The findings showed that varying the amplitude for each category and separating the distance between categories had no effect on the pre-training performance. Therefore, for simplicity, we fixed the amplitude at 0.5 for the baseline configuration of VisualAtom. In our experiments, setting the amplitude to 0.5 produced the best performance \cRio{on} all downstream-tasks. \cRio{Regarding} the performance degradation \cRio{when} setting the amplitude to 1.0, this may be due to the large amplitudes causing the contours to overlap with each other.}

\cRio{
\noindent {\textbf{Best Practice in VisualAtom Pre-training.}}
\label{VisualAtom-best_practice}
According to the results of the experiments above plus some additional experiments not shown here for tuning the parameter combinations, the highest accuracy occurred in category (1000/21,000), instance (1,000), number of orbits ({1, 2, $\cdots$, 200}), vertical and horizontal diameters of orbits ([1.0, 400.0]), orbits spacing (1.0), frequency ({0, 1, $\cdots$, 20}), amplitude (0.5), quantization ({200, 201, $\cdots$,1000}), noise level ([0.0, 1.0]), line-thickness (1), line-color ([0.0, 1.0]), nucleus position ([-1.0, 1.0]$^{2}$), image size (512 × 512) and training epochs (300/90). This best combination of parameters (baseline parameters) is shown in Table~\ref{tableA}. The fine-tuning accuracy is shown in Table~\ref{tab:VisualAtom-1k_baseline} and Table~\ref{tab:comparison_sl_ssl_fdsl}.
Note that we \cRio{explored a much wider range of parameters than those mentioned in this paper, and the accuracy was robust to variations in these parameters.}
}

\begin{table*}[t]
    \begin{center}
    \caption{Comparison of pre-training methods. \cSora{Best and second-best scores at ViT-Tiny are shown in underlined bold and bold, respectively. At ViT-Base, best scores are shown in bold.}}
    \vspace{-8pt}
    \begin{tabular}{l|lcccccccc|c} \toprule[0.8pt]
        Model & Pre-training & Type & C10 & C100 & Cars & Flowers & VOC12 & P30 & IN100 & Average \\\midrule[0.5pt]
        ViT-T & Scratch & -- &  78.3 & 57.7 & 11.6 & 77.1 & 64.8 & 75.7 & 73.2 & 62.6 \\
        & ImageNet-1k & SL &  \underline{\textbf{98.0}} &  \underline{\textbf{85.5}} &  \underline{\textbf{89.9}} &  \underline{\textbf{99.4}} &  \underline{\textbf{88.7}} & \textbf{80.0} & -- & --  \\
        \cmidrule[0.5pt]{2-11}
        
        & ImageNet-1k & SSL (DINO) & \textbf{97.7} & 82.4 & 88.0 & 98.5 & 74.7 & 78.4 & 89.0 & 86.9 \\
        & PASS & SSL (DINO) & 97.5 & 84.0 & 86.4 & 98.6 & \textbf{82.9} & 79.0 & 82.9 & \textbf{87.8} \\
        \cmidrule[0.5pt]{2-11}

        & FractalDB-1k & FDSL & 96.8 & 81.6 & 86.0 & 98.3 & 80.6 & 78.4 & 88.3 & 87.1 \\
        & ExFractalDB-1k & FDSL & 97.2 & 81.8 & 87.0 & \textbf{98.9} & 80.6 & 78.0 & 88.1 & 87.4 \\
        & RCDB-1k & FDSL & 97.0 & 82.2 & 86.5 & \textbf{98.9} & 80.9 & 79.7 & 88.5 & 87.6 \\
        \rowcolor[gray]{0.8} & VisualAtom-1k (ours) & FDSL & 97.6 & \textbf{84.9} & \textbf{88.8} & \textbf{98.9} & 82.0 &  \underline{\textbf{81.2}} &  \underline{\textbf{90.3}} & \underline{\textbf{89.1}} \\\midrule[0.5pt]
        ViT-B
        & RCDB-21k & FDSL & 96.8 & 82.9 & 85.9 & \textbf{99.0} & 81.2 & 81.2 & 90.2 & 88.2 \\
        \rowcolor[gray]{0.8} & VisualAtom-21k (ours) & FDSL & {\bf 97.7} & \textbf{86.7} & \textbf{89.2} & \textbf{99.0} & \textbf{82.4} & \textbf{81.6} & \textbf{91.3} & \textbf{89.7} \\
        \midrule[0.8pt]
    \end{tabular}
    \label{tab:comparison_sl_ssl_fdsl}
    \end{center}
    \vspace{-20pt}
\end{table*}

\begin{table}[t]
    \begin{center}
	\caption{Comparison with fine-tuning accuracy on ImageNet-1k. Res. indicates the image resolution at fine-tuning. Best scores at each dataset scale are shown in bold.}
    \vspace{-10pt}
    \begin{tabular}{lcccc} \toprule[0.8pt]
        Pre-training & Res. & Type & ViT-T & ViT-B \\\midrule[0.5pt]
        Scratch & 224$^{2}$ & -- & 72.6 & 79.8 \\
        ImageNet-21k & 224$^{2}$ & SL & \textbf{74.1} & 81.8 \\
        ExFractalDB-1k & 224$^{2}$ & FDSL & 73.7 & 80.4 \\
        ExFractalDB-21k & 224$^{2}$ & FDSL & 73.6 & \underline{\textbf{82.7}} \\
        RCDB-1k & 224$^{2}$ & FDSL & 73.1 & 82.3 \\
        RCDB-21k & 224$^{2}$ & FDSL & 72.8 & 82.4 \\
        \rowcolor[gray]{0.8} VisualAtom-1k (ours) & 224$^{2}$ & FDSL & \underline{\textbf{74.2}} & 82.3 \\
        \rowcolor[gray]{0.8} VisualAtom-21k (ours) & 224$^{2}$ & FDSL & 73.8 & \underline{\textbf{82.7}} \\\midrule[0.5pt]
        ImageNet-21k & 384$^{2}$ & SL & - & 83.0 \\
        JFT-300M & 384$^{2}$ & SL & - & \underline{\textbf{84.2}}\\
        \rowcolor[gray]{0.8} VisualAtom-21k (ours) & 384$^{2}$ & FDSL & - & \textbf{83.7} \\
        \bottomrule[0.8pt]
    \end{tabular}
    \label{tab:comparison_imagenet1k}
    \end{center}
    \vspace{-5pt}
\end{table}
\tableC
\subsection{Comparisons}

For comparison with previous works, we use seven datasets for evaluation: Stanford Cars (Cars)~\cite{krause20133d}, Flowers~\cite{nilsback2008automated}, Pascal VOC 2012 (VOC12)~\cite{everingham2015pascal}, Places 30 (P30)~\cite{Kataoka_2020_ACCV}, C10~\cite{krizhevsky2009learning}, C100~\cite{krizhevsky2009learning} and IN100~\cite{Kataoka_2020_ACCV}.
To compare the performance of ViT-B pre-trained on JFT-300M in the original ViT paper \cite{dosovitskiy2021an}, we also report results of ViT-B for the fine-tuning performed using ImageNet-1k~\cite{deng2009imagenet} with resolution of $384 \times 384$ [pixel].

\label{sec:comparision}
\noindent \textbf{Comparison with Previous Works (Table~\ref{tab:comparison_sl_ssl_fdsl}).}
In Table~\ref{tab:comparison_sl_ssl_fdsl}, we show the fine-tuning accuracies obtained using the seven datasets.
Here, the following two sets of experiments are conducted: a one-million image-scale experiment with ViT-T and a ten-million image-scale experiment with ViT-B.

In the \textbf{one-million image-scale experiment with ViT-T}, the following pre-training methods were compared:
\cSora{1) SL with ImageNet-1k,}
2) SSL (DINO~\cite{caron2021emerging}) with \cHiro{ImageNet-1k/PASS}~\cite{asano2021pass},
3) FDSL with FractalDB-1k~\cite{Kataoka_2020_ACCV}, ExFractalDB-1k~\cite{kataoka2022replacing}, RCDB-1k~\cite{kataoka2022replacing}, and VisualAtom-1k. 

The results in Table~\ref{tab:comparison_sl_ssl_fdsl} show that VisualAtom-1k achieved the highest average score\footnote{The average accuracy for SL pre-training was not calculated in the previous paper~\cite{kataoka2022replacing}.} for the SSL and FDSL methods. We observed a 1.3 points performance gap between FDSL with VisualAtom-1k (89.1) and SSL with PASS (87.8). This result is surprising given that
VisualAtom-1k contains 1.0M synthetic images
and
PASS contains 1.4M real images.
Moreover, all of the performance rates obtained with VisualAtom-1k were either equal to, or better than, those obtained with the other FDSL datasets.
\cSora{The finding showed that FDSL with VisualAtom-1k partially outperforms SL with ImageNet-1k \cHiro{on P30 (81.2 vs. 80.0).}
}

In the \textbf{ten-million image-scale experiments with ViT-B}, the two FDSL datasets of RCDB and VisualAtom were scaled up to 21,000 categories.
The findings showed that the average accuracy is further improved from 89.1 (VisualAtom-1k) to 89.7 (VisualAtom-21k), and that the \cHiro{score} obtained using VisualAtom-21k was consistently equal to, or better than, that obtained with RCDB-21k.

\noindent \textbf{ImageNet-1k fine-tuning (Table~\ref{tab:comparison_imagenet1k}).}
Table~\ref{tab:comparison_imagenet1k} shows fine-tuning results obtained using ImageNet-1k.
With ViT-T, we see that
VisualAtom-1k surpassed ImageNet-21k
(74.2 v.s. 74.1).
We also see that relatively small-size datasets are effective for effectively pre-training ViT-T with FDSL.
With ViT-B, we see that the scaled-up VisualAtom-21k improved the accuracy from 82.3 to 82.7, which is comparable to that of ExFractalDB-21k and is better than that of ImageNet-21k.
Finally, we conducted an experiment using pixel resolution of $384 \times 384$ in order to compare results with JFT-300M, a massive milestone research dataset that was originally used to pre-train vision transformers.
Although the VisualAtom-21k did not outperform JFT-300M, it achieved a notable performance with $14.2$ times fewer synthetic images (VisualAtom-21k 83.7 with 21M synthetic images vs. JFT-300M 84.2 with 300M real images).

\noindent \textbf{COCO detection/instance segmentation (Table~\ref{tableC}).}
We also conducted validation experiments with object detection and instance segmentation using the COCO~\cite{lin2014microsoft} dataset. We used Swin Transformer (Base)~\cite{liu2021swin} for the backbone model and Mask R-CNN~\cite{he2017mask} for the head. We pre-trained Swin-Base on VisualAtom-1k/21k for 300/90 epochs and performed fine-tuning using the COCO dataset for 60 epochs. Table~\ref{tableC} shows a comparison of our VisualAtom with previous SL 
and FDSL methods, based on AP scores for detection and instance segmentation tasks.
In the fine-tuning with detection and instance segmentation tasks, the results showed that VisualAtom has a higher pre-training effect than RCDB and slightly lower pre-training effect than ExFractalDB. This order was also observed when the dataset category size were scaled up to 21k categories.

\subsection{Discussion and limitations}
\label{sec:discussion}

\cSora{
\noindent \textbf{What should we focus on in FDSL?} In Section~\ref{Pre-training-experiments-on-contours}, we conducted \cRio{systematic} experiments to investigate the relationship between various contour features and the pre-training effect, \cRio{by} changing \cRio{the} parameters of the visual atomic renderer to control contour features of visual atoms.
We found that extending \cRio{the} variation \cRio{of the following parameters enhances the pre-training effect;} frequency (controlling the shape of contours), number of orbits (controlling the size of contours) and quantization (controlling the smoothness of contours).
In particular, increasing the \cRio{variation} of frequency, in other words, using various contour shapes, greatly improved the pre-training effect.
\cRio{With the insight we gained from this systematic investigation, we are now able to construct better FDSL datasets.
The ability to continuously improve the quality of the images and not only the quantity is what distinguishes FDSL from SL with real images.}
}

\noindent \textbf{Potential of \cSora{VisualAtom as a }FDSL dataset.} \cSora{In Section~\ref{sec:comparision}, we validated the pre-training effect of FDSL using our proposed VisualAtom by comparing with many other benchmarks.}
We found that VisualAtom enhances the pre-training effect for relatively small datasets. In particular, fine-tuning comparisons of ViT-T pre-trained on VisualAtom-1k outperformed all other previous FDSL datasets, including ExFractalDB-1k and RCDB-1k (e.g., VisualAtom-1k 89.1 vs. RCDB-1k 87.6, with average accuracies shown in Table~\ref{tab:comparison_sl_ssl_fdsl}). The effectiveness of VisualAtom is also seen in Table~\ref{tab:comparison_imagenet1k}. \cRio{The ViT-T pre-trained on VisualAtom-1k is better than ViT-T pre-trained on ImageNet-21k.} Although a balance exists between model size and pre-training dataset \cRio{size}, the proposed method is also more accurate than ViT-T pre-trained using RCDB-1k/ExFractalDB-1k. Moreover, when conditions were aligned with \cHiro{ViT-B} pre-trained using JFT-300M, our proposed VisualAtom-21k differed by only a 0.5 point gap on ImageNet-1k fine-tuning with 1/14 the number of images. These findings indicate that approaches using contour-based FDSL datasets have considerable potential to enhance the pre-training effect of ViT. 

\noindent \textbf{Limitation.} VisualAtom can still be improved. For example, even though one-million image-scale pre-training significantly improved the recognition performance with ViT-T and relatively small down-stream tasks (see Table~\ref{tab:comparison_sl_ssl_fdsl} showing the results obtained using our VisualAtom-1k with ViT-T), ten-million image-scale pre-training cannot surpass the related FDSL dataset (see Table~\ref{tab:comparison_imagenet1k} showing the results obtained using ExFractalDB-21k/VisualAtom-21k pre-trained with ViT-B). At this stage, we \cRio{speculate} that fractal-like, recursively intricate patterns could conceivably increase the complexity of visual atoms and bring about further improvements.
We intend to further explore ways in which complex patterns can be generated efficiently in order to extend the effectiveness of pre-training without using any real images or human supervision.

\section{Conclusion}
\cRio{In this study, we proposed the visual atomic renderer and how it can control contour features of synthesized images from a mixture of sinusoidal waves.
The simplicity and flexibility of the virual atomic renderer allowed us to perform a systematic exploration of the design space of contour-oriented synthetic datasets.
We found that the rich variation of contour shapes among categories enhances the pre-training effect.
We als found that our proposed VisualAtom dataset outperforms all existing FDSL datasets.}
In a fine-tuning experiment using ImageNet-1k, FDSL with VisualAtom-21k achieved a top-1 accuracy of 83.7\%, and closed the gap in accuracy between FDSL and SL on JFT-300M to 0.5\%.
In addition, compared to JFT-300M, VisualAtom-21k, \cRio{has only 1/14 the number of images}.

\section*{Acknowledgement}
This paper is based on results obtained from a project, JPNP20006, commissioned by the New Energy and Industrial Technology Development Organization (NEDO). Computational resource of AI Bridging Cloud Infrastructure (ABCI) provided by National Institute of Advanced Industrial Science and Technology (AIST) was used.
We want to thank Junichi Tsujii, Yutaka Satoh, Ryo Nakamura, Ryosuke Yamada, Kodai Nakashima, Xinyu Zhang, Zhaoqing Wang, Toshiki Omi, Seitaro Shinagawa, Koshi Makihara for their helpful comments in research discussions.


\end{document}